\documentclass[runningheads]{llncs}
\usepackage{graphicx}
\usepackage{amsmath,amssymb} 
\usepackage{subcaption}
\usepackage{multirow}

\newcommand{\eg}[1]{e.g.}
\newcommand{\ie}[1]{i.e.}

\begin{document}
%===========================================================

\title{Robust Video Background Identification by Dominant Rigid Motion Estimation\thanks{This work was supported by the Singapore PSF grant 1521200082, and partly done when Kaimo, Nianjuan and Jiangbo were with Advanced Digital Sciences Center (ADSC), Singapore.}} % Replace your paper's title here
\titlerunning{Robust Background Motion Estimation in Dynamic Videos} % Replace an abstracted version of your paper's title here

%===========================================================

\author{Kaimo Lin\inst{1} \and
Nianjuan Jiang\inst{2} \and
Loong Fah Cheong\inst{1} \and
Jiangbo Lu\inst{2} \and 
Xun Xu\inst{1}}
%
%Please include author names in full in the paper, 
%If any authors have names that can be parsed into FirstName LastName in multiple ways, please include the correct parsing, in a comment to the volume editors:
%\index{Lastnames, Firstnames}

\authorrunning{K. Lin et al.} % A shorter version of authors' name

%===========================================================

\institute{National University of Singapore, Singapore \\
\email{linkaimo1990@gmail.com, \{eleclf,elexuxu\}@nus.edu.sg} \and
Shenzhen Cloudream Technology, China \\
\email{\{jiangnj,jiangbo\}@cloudream.com}}

\maketitle

%===========================================================
\begin{abstract}
The ability to identify the static background in videos captured by a 
moving camera is an important pre-requisite for many video applications 
(\eg, video stabilization, stitching, and segmentation). Existing methods 
usually face difficulties when the foreground objects occupy a larger 
area than the background in the image. Many methods also cannot scale up 
to handle densely sampled feature trajectories. In this paper, we propose 
an efficient local-to-global method to identify background, based on 
the assumption that as long as there is sufficient camera motion, the 
cumulative background features will have the largest amount of trajectories. 
Our motion model at the two-frame level is based on the epipolar geometry 
so that there will be no over-segmentation problem, another issue that 
plagues the 2D motion segmentation approach. Foreground objects erroneously 
labelled due to intermittent motions are also taken care of by checking 
their global consistency with the final estimated background motion. Lastly, 
by virtue of its efficiency, our method can deal with densely sampled 
trajectories. It outperforms several state-of-the-art motion segmentation 
methods on public datasets, both quantitatively and qualitatively.
\end{abstract}

%===========================================================
\section{Introduction}
Identifying background features from an image sequence is an important 
vision task, subserving many other applications such as video stabilization, 
3D scene reconstruction, background color model estimation for video 
segmentation, etc.. When the camera is stationary, this task is considerably 
simplified. In this paper, we focus on the difficult scenarios when the 
camera is moving and that the foreground might occupy an image area larger 
than the background. Our objective is to identify those static parts of the 
background (hence forth to be called just background, unless otherwise 
specified) even though they may only occupy a small part of the scene in 
some frames. 

RANSAC is the prevailing method for finding background features across 
two frames, when the assumption that the vast majority of the feature 
matches belong to that of the background is true. For videos with large 
moving foreground, it is evident that this simple strategy will fail 
\cite{Liu2013,meshflow2016}. A recent method \cite{movingECCV16} utilizes 
segmentation from the previous frames to help resolve difficulties 
experienced in subsequent parts of the video. This propagation strategy 
may relieve but not remove the aforementioned problems entirely. For 
instance, it depends on the quality of the previous segmentation: 
foreground motions that are intermittent, i.e., stationary at irregular 
intervals, may be accidentally labelled as background, and this wrong 
label is in turn erroneously propagated to later frames. Trajectory-based 
motion segmentation methods consider the entire trajectories across all 
the frames, and generally do not suffer from this kind of error propagation. 
Among this class of approaches, the 3D motion segmentation methods 
\cite{ma2007,zhuwenICCV,xu2018motion} are usually computationally too 
expensive to process densely sampled feature trajectories. 2D motion 
segmentation methods \cite{FBMS59,motion2d} may be fast but produce 
over-segmented results when the background features exhibit large depth 
variation.

In this paper, we address the background identification problem based on 
two observations that we believe to be true most of the times. Firstly, 
the background should be visible in every frame, even though it may not 
occupy the largest area in these frames. Secondly, as long as there is 
enough camera motion such that there is enough turnover of the background 
features, i.e., enough new background features enter into the field of view, 
collecting all these background features together will usually make them 
the group with the largest number of feature trajectories. Therefore, the 
background identification problem becomes that of linking the new features 
into their proper groups as they enter into view. This is done via some 
features that are visible in both the old and the new frames, but of the 
details, more later. While there might also be turnover of the foreground 
objects, they usually cannot be linked together as one group.

\begin{figure*}[!htb]
\centering
\includegraphics[width=1.0\linewidth]{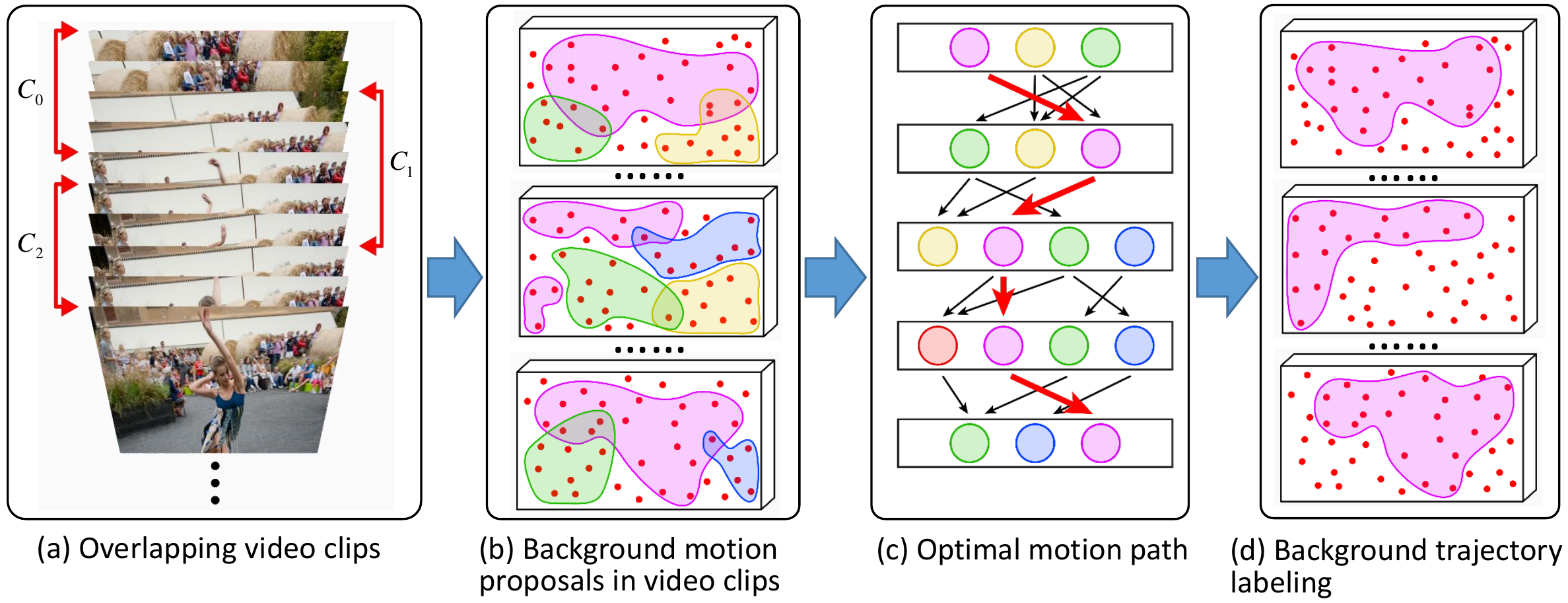}
\caption{Pipeline of our background feature identification method. (a) 
Divide sequence into multiple overlapping clips $\{C_i\}$ Red points in (b) 
and (d) are feature trajectories in video clips. Red arrows in (c) is the 
global dominant rigid motion that contains the largest amount of feature 
trajectories.} 
\label{fig:pipeline}
\vspace{-0.3cm}
\end{figure*}

Since we assume that there is enough camera motion, most trajectories will 
not be visible throughout the entire duration. Thus we divide the video 
into multiple short overlapping clips (Fig.~\ref{fig:pipeline} (a)) and 
from the many potential background motion groups identified from the short 
clips (Fig.~\ref{fig:pipeline} (b)), we attempt to link these groups and 
then identify the linked group that is globally the most dominant, i.e., 
largest. For the linking step, we construct a directed graph with the local 
motion candidates being the nodes, and two nodes from neighboring video 
clips are connected by an edge if they share common trajectories, with the 
weight of the edge being related to the size of the groups involved. Then 
we search through all the possible motion paths (Fig.~\ref{fig:pipeline} 
(c)) that traverse from the first video clip to the last one. The optimal 
motion path is the one that covers the largest amount of feature trajectories, 
and thus corresponds to the desired background motion. Finally, to deal 
with foreground objects with intermittent motions and thus being labelled 
wrongly as background during the frames when they are not moving, we 
enforce global consistency in the trajectory labeling to rectify the errors 
(Fig.~\ref{fig:pipeline} (d)). 

To describe the motions in the local clips, we use a set of epipolar 
geometries (\textbf{EG}) computed from neighboring video frames. Therefore, 
our method can effectively handle videos with large depth variation. 
Our method is much simpler compared to the existing motion segmentation 
techniques, since we are only concerned with the background motion and 
treat all other motions as foreground motions. Our motion estimation method 
is very fast, and the whole pipeline can run very efficiently even with 
densely sampled trajectories.

%===========================================================
\section{Related Work}
While the aim of video object segmentation is to segment out the foreground 
objects from a video sequence, this approach can indeed be used to perform 
background extraction. However, many of these works require some degree of 
human intervention, such as the semi-supervised methods  
\cite{CVPR10_Label_Propagation,activeeccv2012,voxelECCV,seamCVPR2014,Perazzi_2015_ICCV,Maerki_CVPR_2016} which require a small amount of manual annotation, 
and the fully-supervised methods \cite{snapcut2009,disc2012,jumpcut2015} 
which require repeating result correction by the user. Our work is completely 
autonomous, like the unsupervised methods of video object segmentation 
\cite{BMVC2014,FliICCV2013,FastICCV2013,SaliencyCVPR,CasualCVPR,ObjectCVPR,Lee2011}. However, the underlying assumption of our work is less brittle compared 
to those subscribed to by these works. For instance, \cite{Lee2011,ObjectCVPR} 
rely on the ability of object proposals to detect the foreground objects; 
this might fail to work when the foreground objects have complex non-compact 
shapes. \cite{FastICCV2013,SaliencyCVPR} detect foreground objects by 
analysing the 2D motion field based on a simple assumption that they usually 
move differently from their surroundings. However, the 2D motions of some 
background objects may also have this property if the background has large 
depth variation. 

Trajectory-based motion segmentation methods usually produce multiple 
independent motion groups; however, they usually stop short of actually 
identifying the background motion, or just use a simple background metric 
such as size. Furthermore, due to the computational demand of the 3D motion 
segmentation methods \cite{missing2008,ssc2009,zhuwenICCV,xu2018motion}, 
these works lack the ability to deal with large amount of feature 
trajectories. For long and densely sampled feature trajectories, some fast 
2D methods \cite{FBMS59,motion2d} may be used. However, their results may 
become over-segmented when the figure-ground is complex, for instance, 
scenes with large depth variation. Our method not only processes large 
amount of trajectory in an efficient way, but also handle videos with 
complex scene structures.

In addition to the above approaches, Bideau and Learned-Miller 
\cite{movingECCV16} proposed a probabilistic model to segment all the 
moving objects. Similar to many of the above approaches, their method 
cannot handle scenes with large foreground objects and may face difficulties 
when dealing with intermittent motions. Zhang \textit{et al.}'s work 
\cite{robust2016} built a directed graph from local motion groups; this 
approach is closely related to ours. Our method is different from theirs 
in two aspects. Most importantly, we identify the background motion as the 
one with the largest amount of feature trajectories in the aggregate, while 
their method is based on the conjecture that the background trajectory matrix 
should exhibit a lower rank. This assumption is a serious qualification: it 
will not work if the foreground motions are also rigid. Another difference 
is that our method tries to find as many rigid motions as possible in each 
video clip, instead of committing to a clean segmentation for all the feature 
trajectories based on traditional motion segmentation methods. This allows us 
to recover from errors that might arise at the local clip levels.

%===========================================================
\section{Algorithm Overview}
Given a dynamic video with large foreground objects, we first extract 
feature trajectories $\left\{T_0,...,T_i\right\}$ using the method 
\cite{dense2010}. We followed the instruction from \cite{dense2010} 
and used the author’s codes to generate the feature trajectories. 
Trajectories are extracted at a fixed interval in both horizontal and 
vertical directions. Our algorithm takes these trajectories as input and 
outputs those belong to the static background by estimating the dominant 
rigid motion in the video. The system pipeline is shown in 
Fig.~\ref{fig:pipeline}. 

The system first divides the input video into many short overlapping 
video clips of variable lengths, i.e.,, $\left\{C_0,...,C_i\right\}$ 
(Fig.~\ref{fig:pipeline} (a)). Inside each video clip $C_i$, we propose 
multiple motion candidates $\{M_i^0,...,M_i^m\}$, which contain different 
sizes of trajectory groups (Fig.~\ref{fig:pipeline} (b)). Each $M_i^m$ 
represents a rigid motion inside $C_i$. The global background motion for 
the entire video is a collection of the background motions in these video 
clips, i.e., select one $M_i^m$ for each $C_i$. 

Since the background motions inside the video clips may not always 
be the majority one with the presence of large foreground objects, 
we identify it using a graph search method in a global manner 
(Fig.~\ref{fig:pipeline} (c)). Specifically, we construct a directed 
graph with $M_i^m$ in each $C_i$ as graph nodes. Directed edges are only 
created from nodes in $C_i$ to those in $C_{i+1}$ with shared trajectories. 
Among all the possible motion paths that start from the very first clip 
to the last one, we select the one that contains the largest amount of 
feature trajectories as our global background motion.

For objects with intermittent motions, \ie, they may be static in 
some video clips, their motions in those clips may be wrongly labeled as 
background. Therefore, we further perform a temporal consistency check 
and background motion refinement to rectify those wrongly labeled 
trajectories (Fig.~\ref{fig:pipeline} (d)).

%===========================================================
\section{Motion Estimation in Video Clips}
\label{sec:localmotion}
\begin{figure}
\centering
\includegraphics[width=0.65\linewidth]{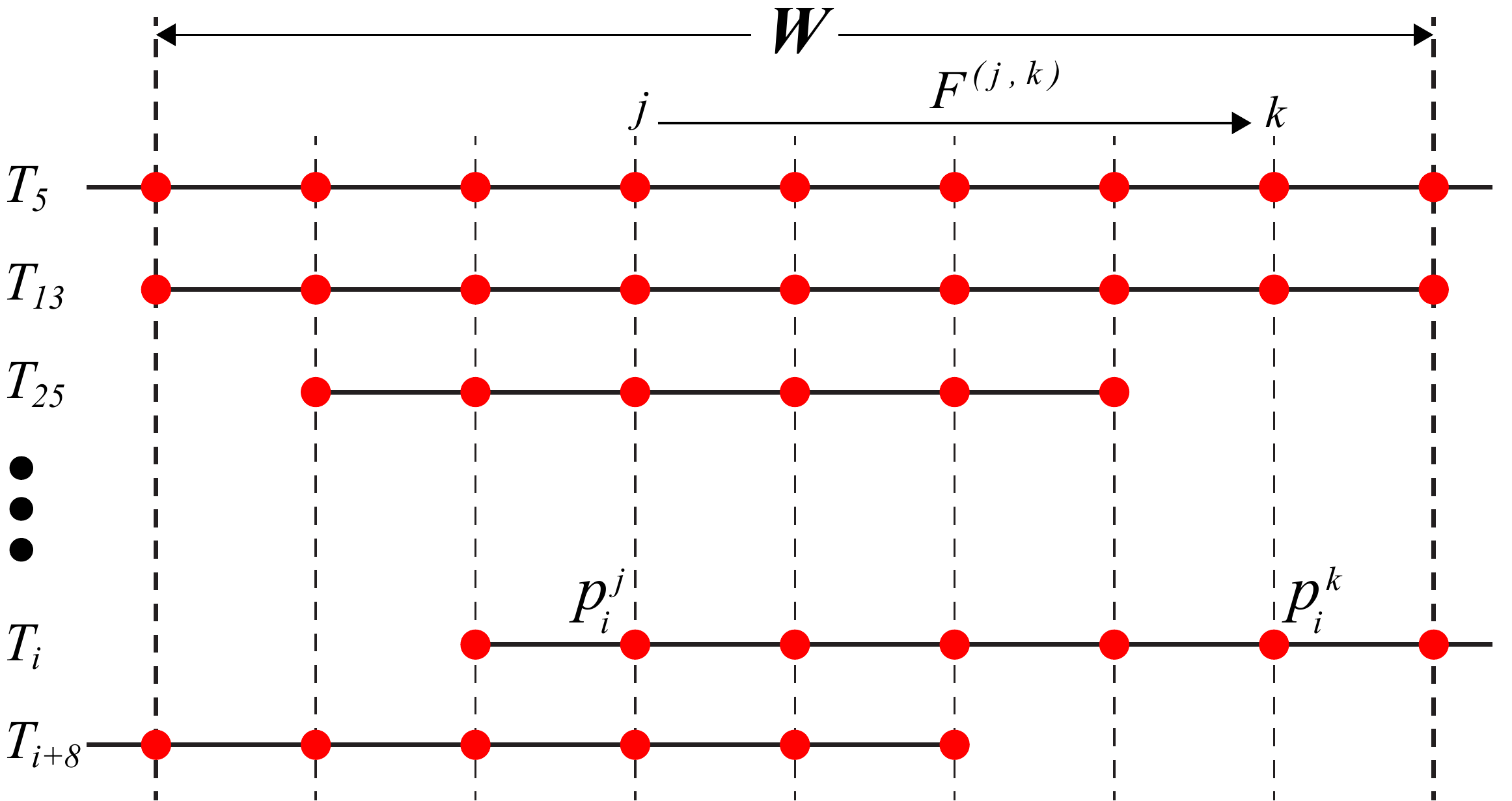}
\caption{Motion model defined by a set of feature trajectories inside a 
clip $C$ with time window $W$. Red points are tracked features on each 
trajectory.} 
\label{fig:Tvisible}
\vspace{-0.4cm}
\end{figure}

Given a clip $C$ with time window $W$, we divide the feature trajectories 
into three categories, i.e., $\left\{T_i^t\right\}_{W}$. The indicator $t$ 
is set according to the following rules. 

\begin{equation}
t=\begin{cases}
1,\quad \text{$T_i$ is always visible within $W$ (e.g., $T_5$ and $T_{13}$ in Fig.\ref{fig:Tvisible}})\\
0,\quad \text{$T_i$ is partially visible within $W$ (e.g., $T_{25}$ in Fig.\ref{fig:Tvisible})}\\
-1,\quad \text{$T_i$ is invisible within $W$}
\end{cases}
\end{equation}\label{eq:Tvisible}

The indicator $t$ of a trajectory $T_i$ is set to $1$ if the trajectory 
lives throughout the entire time window (we call them full-length 
trajectories inside the time window), otherwise, it is set to $0$ or $-1$. 
This labeling of feature trajectories will be used to facilitate dividing 
the input video into overlapping video clips. 

%===========================================================
\subsection{Video Clips Generation}
Starting from the first video frame, we keep adding new frames to expand 
the time window $W_0$ of the first video clip $C_0$ as long as the number 
of the full-length feature trajectories inside $C_0$, \ie, 
$\left\{T_i^t\right\}_{C_0}^{t=1}$, is more than $80\%$ of all the visible 
trajectories inside it, as define below,
\begin{equation}
\frac{|\{T_i^t\}_{C_0}^{t=1}|}{|\{T_i^t\}_{C_0}^{t=1}|+|\{T_i^t\}_{C_0}^{t=0}|}\geq 0.8
\end{equation}
The next video clip $C_{1}$ starts from the middle frame of $C_0$ and 
expands in the same way. We repeat this process until the end of the video sequence is reached. 

We experimentally set the $80\%$ ratio here to achieve a balance between 
enough camera movement within a clip and adequate trajectory overlap with 
the next video clip. As a result, long video clips will be generated when 
the camera movement is small. This ensures that even under small camera 
movement, there will always be enough motion cues within a particular clip 
for robust motion estimation. 

\begin{figure*}
\centering
\begin{tabular}{@{} cc @{}}
	\begin{subfigure}{.35\textwidth}
  		\centering
  		\includegraphics[width=1.0\linewidth]{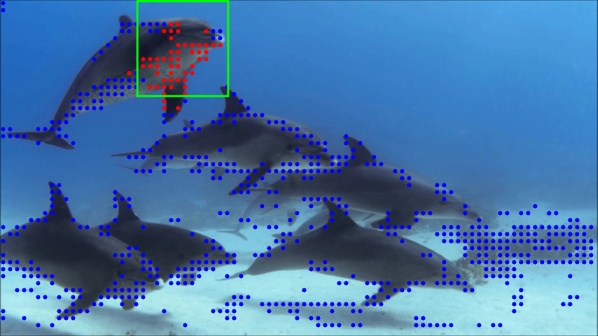}
	\end{subfigure} \quad
	\begin{subfigure}{.35\textwidth}
  		\centering
  		\includegraphics[width=1.0\linewidth]{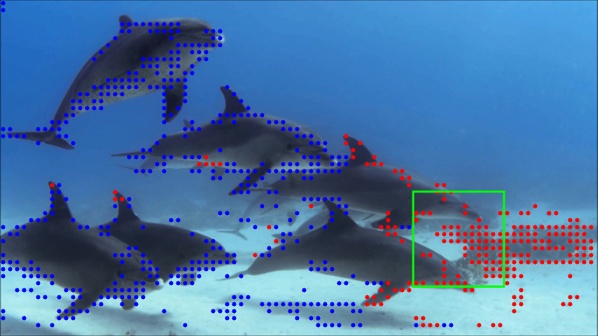}
	\end{subfigure} \\
	\begin{subfigure}{.35\textwidth}
  		\centering
  		\includegraphics[width=1.0\linewidth]{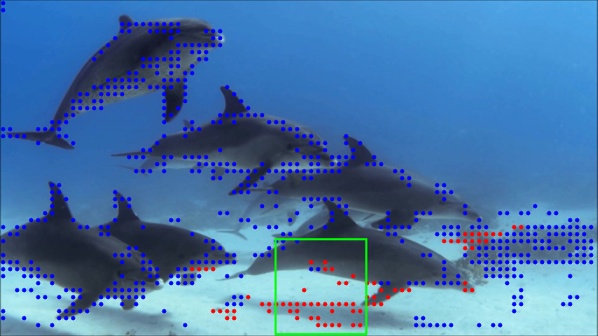}
	\end{subfigure} \quad
	\begin{subfigure}{.35\textwidth}
  		\centering
  		\includegraphics[width=1.0\linewidth]{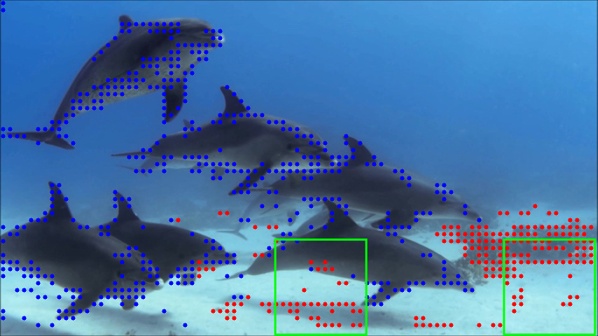}
	\end{subfigure}
\end{tabular}
\caption{Examples of motion candidates in a video clip. Full-length 
trajectories inside the green rectangles are used to estimate the 
best-fitting motion models. Red points represent feature trajectories 
that are labeled as members of the motion candidates, while blue points 
are non-member feature trajectories.} 
\label{fig:local}
\vspace{-0.3cm}
\end{figure*} 

%===========================================================
\subsection{Motion Model}
\label{sec:model}
In our method, a rigid motion $M$ defined by a set of trajectories 
$\left\{T_i^t\right\}_{C}^{t=0,1}$ is represented by a series of 
fundamental matrices $\left\{F^{(j,k)}\right\}_C^{0<|j-k|\leq r}$, 
where $j$, $k$ are the video frame indices inside clip $C$ (see 
Fig.~\ref{fig:Tvisible}). Here, $r$ is empirically set to five as the 
feature tracking error is usually acceptable within this range for 
good fundamental matrix estimation. We estimate the fundamental matrix 
$F^{(j,k)}$ by applying the 8-point algorithm \cite{Hartley2004} on the 
feature matches extracted from these trajectories that are both visible 
at frames $j$ and $k$. 

To decide if a trajectory $T_i$ belongs to a known rigid motion 
$\left\{F^{(j,k)}\right\}_C^{0<|j-k|\leq r}$, we compute its geometric 
errors based on the point-to-epipolar-line distance. Specifically, for 
each $F^{(j,k)}$, if $T_i$ is both visible at frames $j$ and $k$, we 
extract a feature match $(p_i^j,p_i^k)$ from it (see Fig.~\ref{fig:Tvisible}) 
and compute its geometric error with respect to $F^{(j,k)}$ as follows:  
\begin{equation}
\label{eqn:geo_error}
  g(p_i^j, p_i^k, F^{(j,k)}) = d(p_i^k,F^{(j,k)} p_i^j), 
\end{equation}
where $p_i^j$ and $p_i^k$ are all homogeneous coordinates and $d$ denotes 
the Euclidean point-to-line distance. If the error is smaller than a 
threshold $\varepsilon_f$ ($\varepsilon_f = 1.5$ in our implementation), 
we mark this feature match as a positive match. If more than $90\%$ of the 
tested feature matches from $T_i$ are marked as positive, we label $T_i$ as 
a member of the rigid motion.

%===========================================================
\subsection{Motion Estimation in Video Clips}
\label{sec:localmotion}
Knowing that the background features may not constitute the majority in a 
video clip, we cannot hypothesize the background motion in the regular 
RANSAC fashion, \ie, randomly select eight full-length trajectories in 
RANSAC iterations. Instead, we adopt a local scanning scheme to effectively 
detect multiple motion candidates inside each clip. 

For each video clip $C_i$, we first divide the starting frame of $C_i$ into 
overlapping cells, \eg, $30\%$ overlap with a uniform size $L\times L$. For 
each cell, we collect the full-length trajectories inside it, and estimate 
a best-fitting motion in the regular RANSAC fashion. Specifically, in each 
RANSAC iteration, we randomly select eight full-length trajectories to 
compute a motion as described in Sec.~\ref{sec:model}. Then, we label the 
rest of the full-length trajectories inside the cell and count the inliers. 
If the inliers of the final best-fitting motion exceeds $80\%$ of all the 
full-length trajectories within that cell, we regard the estimated motion 
$M_i^m$ as a plausible background motion candidate for $C_i$. Otherwise, it 
is discard. Note that we only use full-length trajectories here to ensure 
that we can compute a motion with a minimum eight trajectories.

After the RANSAC model fitting for all the single cells, we obtain a set of 
motion candidates $\left\{M_i^m\right\}$. Since the background region may 
consist of several discrete parts across the image domain due to the large 
dynamic foreground motions and the estimated motions from single cells can 
be locally biased, we also perform the model fitting process on combinations 
of the single cells that yield $\left\{M_i^m\right\}$. In our experiments, 
we find that combinations of up to three single cells are usually sufficient 
for robust background motion estimation. Eventually, successful motions from 
these combined cells together with those from the single cells form the final 
motion candidate set for $C_i$

These motion candidates are estimated using full-length trajectories 
inside local cells. We still need to decide the membership of other 
trajectories inside the clip with respect to each of motion candidates. 
Fig.~\ref{fig:local} shows some labeling results of the motion candidates. 
As we can see, unlike traditional motion segmentation methods that perform 
a clean segmentation of the trajectories and assign a unique label to each 
$T_i$, our method usually generates multiple overlapping motion groups. 
Most importantly, since the motion candidates are estimated from densely 
overlapping cells, there is a high chance that the background will be the 
majority in at least one of these cells, and thus the true background motion 
can be estimated.

%===========================================================
\section{Dominant Motion Path Estimation}
\label{sec:motionpath}
Once we obtain the motion candidates in each video clip, we construct 
a directed graph with these motion candidates as graph nodes (see 
Fig.~\ref{fig:pipeline} (c)). For two neighboring video clips $C_i$ and 
$C_{i+1}$, we create a directed edge between two motion candidates if 
there exist some common trajectories between them, \eg, $M_i^j$ and $M_i^n$ 
in $C_i$ are both connected to $M_{i+1}^k$ in $C_{i+1}$.

%===========================================================
\subsection{Graph Edge Weight}
Among the multiple motion paths that traverse from the first video clip 
to the last, we now seek the optimal one that has the largest sum of 
trajectories along its path, \ie, the dominant rigid motion. A feature 
trajectory $T_i$ may live through multiple video clips and it may not 
always be the member of the motions along the path. For better 
trajectory counting along a motion path, we divide each $T_i$ into $N$ 
sub-trajectories, where $N$ is the number of video clips it spans. We 
then assign each sub-trajectory with a value of $v=\frac{1}{N}$. With 
this simple normalization, we are now ready to count the number of 
trajectories in the following manner. 

\begin{figure}
\centering
\includegraphics[width=0.75\linewidth]{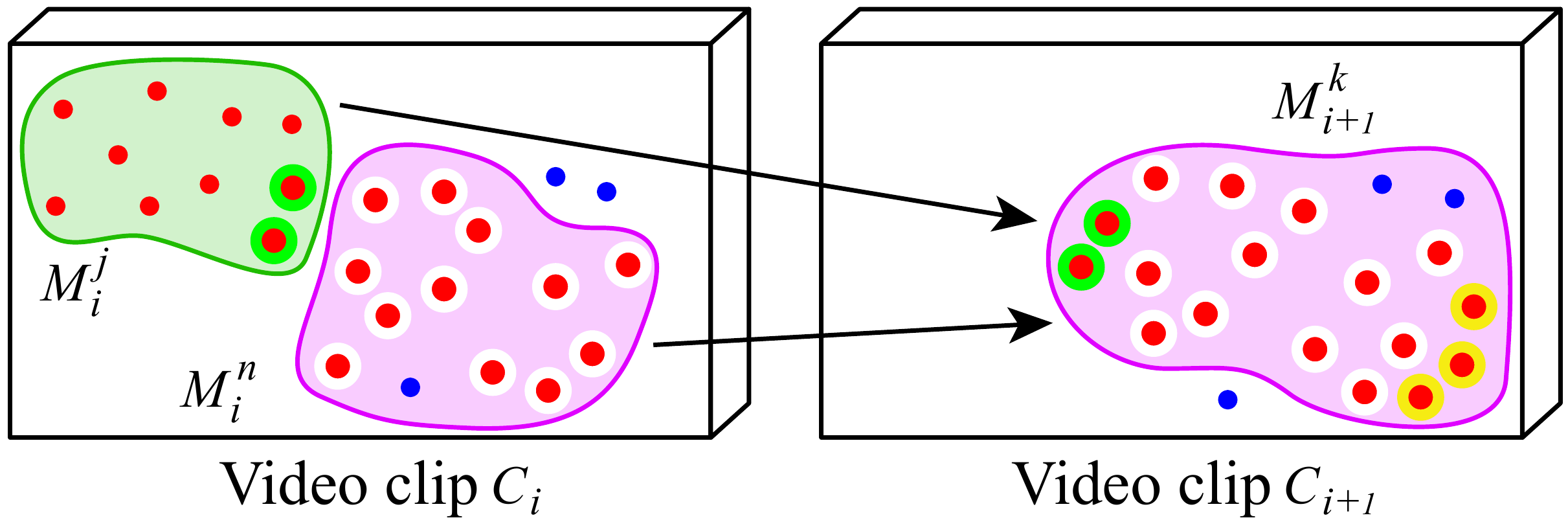}
\caption{Trajectory counting between two connected motions. Green circles: 
shared trajectories between $M_i^j$ and $M_{i+1}^k$. White circles: shared 
trajectories between $M_i^n$ and $M_{i+1}^k$. Yellow circles: newly appeared 
trajectories in $C_{i+1}$ that belong to $M_{i+1}^k$.} 
\label{fig:edge}
\vspace{-0.3cm}
\end{figure}

For an edge between $M_i^j$ and $M_{i+1}^k$, an edge weight is defined as
\begin{equation}
    e_{i, i+1}^{j, k} = \sum_c G(T_c,M_{i+1}^k)\cdot v_c + \sum_n G(T_n,M_{i+1}^k)\cdot v_n,
\label{eqn:connection}
\end{equation}
where $v_c$ are the values of sub-trajectories common to both $M_i^j$ and 
$M_{i+1}^k$, and $v_n$ are the values of those in $M_{i+1}^k$ that have 
newly appeared in $C_{i+1}$. Therefore, we favor those strong connections, 
\eg, edge between $M_i^n$ and $M_{i+1}^k$ in Fig.~\ref{fig:edge} that have 
larger number of shared trajectories, because a consistent background motion 
path should have maximum overlapping trajectory groups. 

$G(T_i,M)$ is a weighting term that reflects the geometric error of a 
sub-trajectory under a certain motion candidate. It is defined as: 
\begin{equation}
    G(T_i,M) = exp\left(-\frac{g_M^i \cdot g_M^i}{2\sigma^2}\right),
\end{equation}
where $\sigma = 0.15$, and $g_M^i$ is the average geometric error of 
trajectory $T_i$ under motion $M$ (see Eq.~(\ref{eqn:geo_error})). When a 
motion candidate is connected to multiple motions with similar number of 
shared trajectories, it will favor the connection with lower fitting error.

%===========================================================
\subsection{Optimal Path Search}
Given a starting motion candidate $M_0^j$ in the first video clip and 
an ending one $M_q^k$ in $C_q$, we estimate the optimal motion path from 
$M_0^j$ to $M_q^k$ that contains the largest number of trajectories by 
defining an objective function that maximizes the edge weights:
\begin{equation}
P(M_0^j, M_q^k) = \max_{(n, k)\in\Phi}\left\{P(M_0^j, M_{q-1}^n) + e_{q-1,q}^{n,k}\right\},
\end{equation}
where $\Phi$ is the set of connected edges from all motion candidates in 
$C_{q-1}$ to $M_{q}^k$. When $q = 1$, we have $P(M_0^j, M_1^k) = e_{0,1}^{j,k} + \Omega(M_0^j)$, where $\Omega(M_0^j)$ is the sum of the sub-trajectories' 
values inside $M_0^j$, weighted by their corresponding geometric errors as 
in Eq.~(\ref{eqn:connection}). 

This optimal motion path searching problem between a motion candidate $M_0^m$ 
in $C_0$ to any other candidate $M_i^m$ in $C_i$ can be easily solved by 
dynamic programming. Finally, the global optimal motion path, starting from 
the first video clip to the last one, is selected by the following:
\begin{equation}
P_{dom} = \max_{j,k}\left\{P(M_0^j, M_{S-1}^k)\right\},
\end{equation}
where $S$ is the total number of the video clips. 

%===========================================================
\section{Background Trajectory Labeling}
After obtaining the optimal motion path $P_{dom}$, \ie, background motion 
group, we label the sub-trajectories inside the selected motion candidates 
as background. Since a feature trajectory usually lives through multiple 
clips, and the optimal motion path may give its sub-trajectories different 
labels due to reasons like intermittent motion of the foreground objects or 
motion estimation errors, we need to perform a temporal consistency check 
for all the trajectories and use only reliable ones to compute the motion 
model of the global background motion. 

Specifically, given the background motion path, we first identify reliable 
background feature trajectories that are entirely covered by the path, 
\ie, those whose all sub-trajectories are labelled as background. Then, 
we compute the global background motion model $M_{back}$, using those 
reliable background trajectories as described in Sec.~\ref{sec:model}. 
Note that the time window of $M_{back}$ covers the entire video. Finally, 
we use $M_{back}$ to label all the feature trajectories in the video. 
And the labelled background trajectories will be the initial output of 
our algorithm.

%===========================================================
\subsection{Local Trajectory Label Filtering}
Due to tracking errors, some background feature trajectories, especially 
those close to object boundaries, may be wrongly labeled as non-background 
trajectories, while most of its neighbors are correctly labeled. To obtain 
spatially-smooth but edge-preserving labeling results, we apply a local 
filtering process on the labeled trajectories based on their color 
similarity and spatial-temporal connectivity. Specifically, we regard two 
trajectories $T_i$ and $T_j$ as neighbors in the spatial-temporal domain 
if they have at least one shared video frame and their minimum distance 
inside these shared frames is less than a threshold ($5\%$ of the frame 
width in our implementation). For each pair of such trajectory neighbors, 
we assign a weight $w_{i,j}$ to them, defined as follows:
\begin{equation}
  w_{i,j} = exp\left(-\frac{d_s^2}{2 \sigma_d^2}\right) \cdot exp\left(-\frac{d_c^2}{2 \sigma_c^2}\right), 
\label{eqn:weight}
\end{equation}
where $d_s$ is the maximum distance between $T_i$ and $T_j$ inside their 
shared frames and $d_c$ is the average RGB color difference of them 
respectively. We set $\sigma_d$ to $2\%$ of the frame width and $\sigma_c = 0.18$ in all our experiments. 

If a feature trajectory $T_i$ is labeled as background initially, we assign 
a value $L(T_i) = 1.0$ to it. Otherwise, $L(T_i) = 0$. The new label of 
$T_i$ is then determined by the following local filtering operation:
\begin{equation}
  L(T_i)^* = \sum_{j\in Ne(i)}w_{i,j} \cdot L(T_j), 
\label{eqn:filtering}
\end{equation}
where $Ne(i)$ is the set of trajectory neighbors of $T_i$. If 
$L(T_i)^* > 0.5$, we set the new label of $T_i$ as background. Otherwise, 
it is labeled as non-background trajectory. The filtering step produces 
the final results for background segmentation.

%===========================================================
\begin{figure*}[!htb]
\centering
\includegraphics[width=0.92\linewidth]{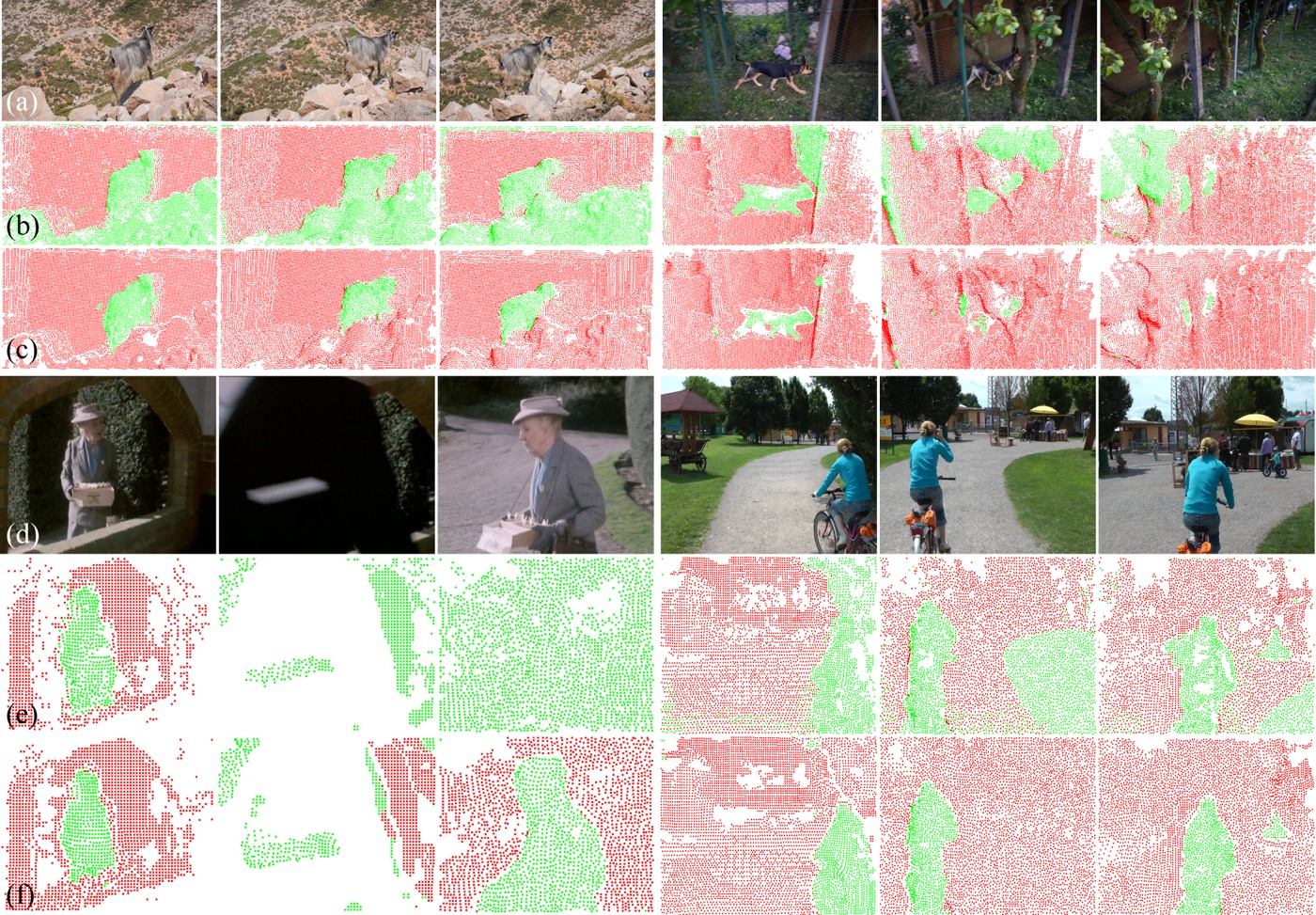}
\caption{Visual comparison with Keuper \textit{et al.} \cite{motion2d}. 
Rows (b) and (e) are results from \cite{motion2d}. Rows (c) and (f) are our 
results. The background features are shown in red and the non-background 
features are in green.} 
\label{fig:iccv}
\end{figure*}

%===========================================================
\section{Experiments}
We compare our method with several most recent state-of-the-art motion 
segmentation and background identification methods, which include the 
motion trajectory segmentation method \cite{motion2d}, dense binary motion 
segmentation method \cite{movingECCV16}, and the robust background 
identification method by Zhang \textit{et al.} \cite{robust2016}. These 
methods represent different approaches for background identification tasks. 
The experiments are conducted with several well-known public datasets, 
where high quality ground truth foreground masks are provided. Specifically, 
we include \textbf{FBMS-59} \cite{FBMS59} (55 videos used), \textbf{DAVIS} 
\cite{DAVIS} (32 videos used), Complex Background Data Set \cite{complex2013} 
(\textbf{CBDS}), Camouflaged Animals Data Set \cite{movingECCV16} 
(\textbf{CADS}), and the videos used in \cite{robust2016} for our analysis. 
These datasets contain various types of dynamic video (\eg, freely-moving 
camera, large depth variation, small foreground movement, highly dynamic 
scenes). Note that we discard some videos from \textbf{FBMS-59} \cite{FBMS59} 
and \textbf{DAVIS} \cite{DAVIS} that violate the following 
criteria: (1) Available foreground mask should cover all the moving objects 
in the scenes. (2) The static background, regardless of its size, should 
always be visible. (3) No severe lens distortion in the video.

%===========================================================
\setlength{\tabcolsep}{4pt}
\begin{table}[!htb]
\begin{center}
\begin{tabular}{|c|c|c|c|c|c|c|}
\hline
\multirow{2}{*}{Method} & \multicolumn{2}{c|}{Precision (\%)} & \multicolumn{2}{c|}{Recall (\%)} & \multicolumn{2}{c|}{F-score} \\ \cline{2-7} 
& \small{FBMS} & \small{DAVIS} & \small{FBMS} & \small{DAVIS} & \small{FBMS} & \small{DAVIS}\\ \hline
Keuper & 95.5 & 99.5 & 89.9 & 91.8 & 91.8 & 95.4 \\ \hline
Ours   & 95.0 & 99.0 & 98.3 & 98.3 & 96.4 & 98.6 \\ \hline
\end{tabular}
\end{center}
\caption{Comparison of precision and recall on datasets \textbf{FBMS} and \textbf{DAVIS} with Keuper \textit{et al.} \cite{motion2d}.}
\label{table:motion2d}
\vspace{-0.5cm}
\end{table}

%===========================================================
\subsection{Comparison with Keuper \textbf{\textit{et al.}} \cite{motion2d}}
Keuper \textit{et al.}'s method \cite{motion2d} is currently the 
state-of-the-art 2D motion segmentation method. To compare with their 
method, we test both methods on datasets \textbf{FBMS-59} \cite{FBMS59} 
and \textbf{DAVIS} \cite{DAVIS}. The input feature trajectories are 
extracted by \cite{dense2010} for both methods. Since the output of 
\cite{motion2d} are multiple groups of feature trajectories, we select 
the one that contains the largest amount of trajectories and label the 
features inside as background. The comparison of average precision and 
recall for background features identification is shown in 
Tab.~\ref{table:motion2d}. As we can see from the table, our method 
achieves similar high accuracy as Keuper \textit{et al.}'s method, while 
the recall of our method is significantly better. Fig.~\ref{fig:iccv} 
shows some typical cases that may fail Keuper \textit{et al.}'s method. 
Firstly, their method may not work well on scenes with large depth 
variation (top two cases in Fig.~\ref{fig:iccv}), which is a typical 
limitation of 2D motion segmentation methods without using a projective 
motion model. Secondly, when the background is severely occluded by the 
foreground moving objects (bottom left case in Fig.~\ref{fig:iccv}), their 
method may fail to track the background motion, and instead, creates new 
motion groups. Finally, since their method utilizes color information 
along with other motion cues in their energy function, it may over-segment 
the background region if it contains components with large color difference 
(bottom right case in Fig.~\ref{fig:iccv}). Our method, on the other hand, 
can very well handle such difficult cases.

%===========================================================
\begin{figure*}[!htb]
\centering
\setlength{\tabcolsep}{1.0pt}
\begin{tabular}{@{} ccccc @{}}
    & Original frame & Ground truth mask & Result from \cite{movingECCV16}  & Our result 
\\
    \rotatebox[origin=c]{90}{\textit{traffic}} & 	
    \begin{subfigure}{.236\textwidth} 
	    \centering 
	    \includegraphics[width=1.0\linewidth]{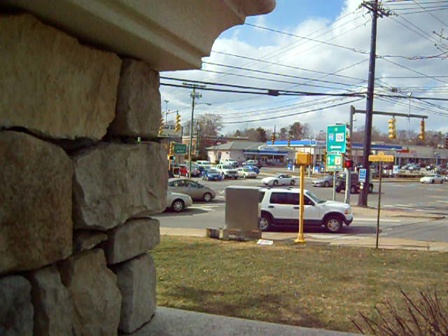} 
	    \vspace{-0.35cm} 
	\end{subfigure} &
    \begin{subfigure}{.236\textwidth} 
	    \centering 
	    \includegraphics[width=1.0\linewidth]{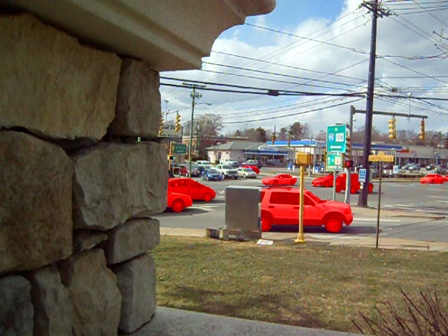} 
	    \vspace{-0.35cm} 
	\end{subfigure} &
    \begin{subfigure}{.236\textwidth} 
	    \centering 
	    \includegraphics[width=1.0\linewidth]{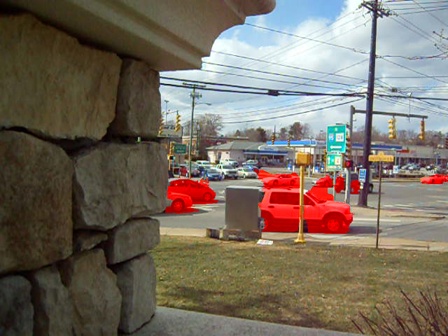} 
	    \vspace{-0.35cm} 
	\end{subfigure} &
    \begin{subfigure}{.236\textwidth} 
	    \centering 
	    \includegraphics[width=1.0\linewidth]{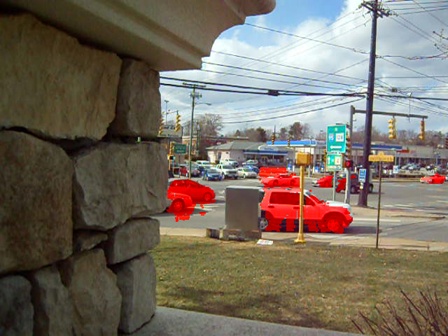} 
	    \vspace{-0.35cm} 
	\end{subfigure} 
\\
    \rotatebox[origin=c]{90}{\textit{chameleon}} & 	
    \begin{subfigure}{.236\textwidth} 
	    \centering 
	    \includegraphics[width=1.0\linewidth]{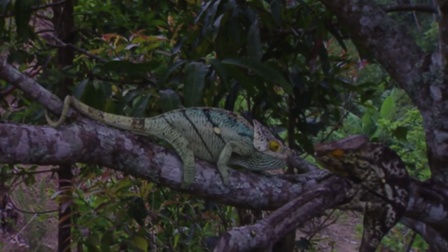} 
	    \vspace{-0.35cm} 
	\end{subfigure} &
    \begin{subfigure}{.236\textwidth} 
	    \centering 
	    \includegraphics[width=1.0\linewidth]{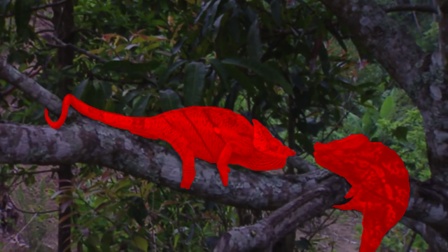} 
	    \vspace{-0.35cm} 
	\end{subfigure} &
    \begin{subfigure}{.236\textwidth} 
	    \centering 
	    \includegraphics[width=1.0\linewidth]{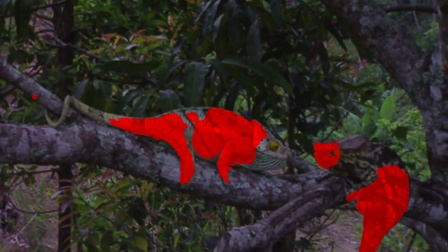} 
	    \vspace{-0.35cm} 
	\end{subfigure} &
    \begin{subfigure}{.236\textwidth} 
	    \centering 
	    \includegraphics[width=1.0\linewidth]{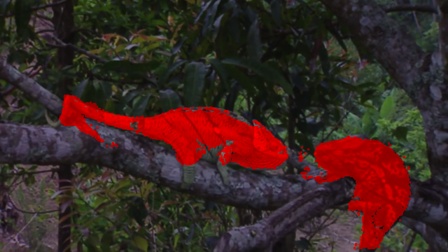} 
	    \vspace{-0.35cm} 
	\end{subfigure} 
\\
    \rotatebox[origin=c]{90}{\textit{scorpion4}} & 	
    \begin{subfigure}{.236\textwidth} 
	    \centering 
	    \includegraphics[width=1.0\linewidth]{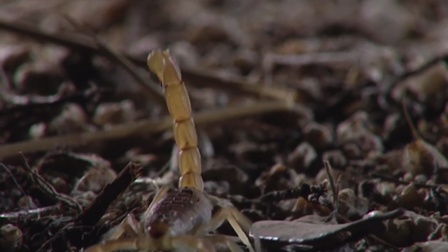} 
	    \vspace{-0.35cm} 
	\end{subfigure} &
    \begin{subfigure}{.236\textwidth} 
	    \centering 
	    \includegraphics[width=1.0\linewidth]{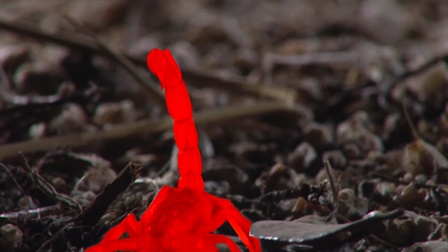} 
	    \vspace{-0.35cm} 
	\end{subfigure} &
    \begin{subfigure}{.236\textwidth} 
	    \centering 
	    \includegraphics[width=1.0\linewidth]{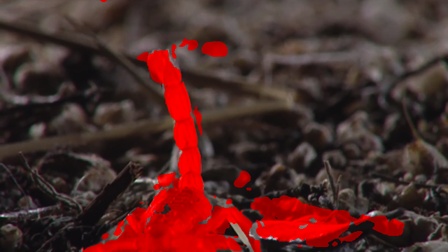} 
	    \vspace{-0.35cm} 
	\end{subfigure} &
    \begin{subfigure}{.236\textwidth} 
	    \centering 
	    \includegraphics[width=1.0\linewidth]{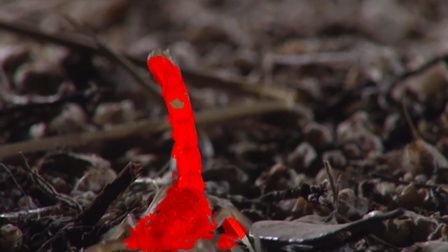} 
	    \vspace{-0.35cm} 
	\end{subfigure} 
\\
\end{tabular}
\caption{Visual comparison with Bideau and Learned-Miller's method \cite{movingECCV16}.}
\label{fig:eccv}
\end{figure*}

%===========================================================
\subsection{Comparison with Bideau and Learned-Miller's method \cite{movingECCV16}}
We also compare our method with Bideau and Learned-Miller's method 
\cite{movingECCV16}, which is a recent foreground object segmentation 
method. The datasets we use in this experiment are the Complex Background 
Data Set from \cite{complex2013} and the Camouflaged Animals Data Set from 
their own work. Since the main purpose of their method is segmenting 
foreground objects and the objects in these datasets are relatively small, 
we compute the precision and recall of pixels/features that locate on the 
foreground moving objects instead of background this time. The results are 
reported in Tab.~\ref{table:eccv}. As we can see, in most cases, our method 
achieves better precision and recall. In general, our method produces more 
accurate and complete segmentation of the foreground objects, see examples 
in Fig.~\ref{fig:eccv}. The method proposed by \cite{movingECCV16}, on the 
other hand, has difficulty dealing with foreground intermittent motions 
(`\textit{chameleon}' sequence in Fig.~\ref{fig:eccv}), which are explicitly 
handled in our method by utilizing global motion cues.

%===========================================================
\subsection{Comparison with Zhang \textbf{\textit{et al.}} \cite{robust2016}} 
To evaluate the performance of our method on videos with large foreground 
objects, we compare our method with Zhang \textit{et al.}'s method 
\cite{robust2016} on the highly dynamic videos used in their work. Some 
segmentation results are shown in Fig.~\ref{fig:sig}. As we can see, our 
method can also produce very good segmentation results on these videos. 
Since we use projective motion model in our motion estimation step, the 
estimated background motion usually covers the entire background region. 
The motion segmentation method by Zhang \textit{et al.} may produce 
over-segmented results as reported in their paper (yellow and green points 
on the background wall in Fig.~\ref{fig:sig} (a)). Therefore, it requires 
further post-processing to merge these background motion groups. Also, the 
background metric used in their method may be violated if the foreground 
motions are also rigid motions or nearly rigid motions (Fig.~\ref{fig:sig} 
(e)). Our background metric is based on the total number of feature  
trajectories a rigid motion contains in the entire video, which is proved 
to be more robust according to our experiment results. 

\setlength{\tabcolsep}{5.5pt}
\begin{table}[!h]
\centering
\label{my-label}
\begin{tabular}{|c|c|c|c|c|c|}
\hline
\multicolumn{2}{|c|}{\multirow{2}{*}{Sequence}} & \multicolumn{2}{c|}{Bideau \cite{movingECCV16}} & \multicolumn{2}{c|}{Ours} \\ \cline{3-6} 
\multicolumn{2}{|c|}{}      & preci. & recall & preci. & recall \\ \hline
\multirow{5}{*}{\rotatebox[origin=c]{90}{\textbf{CBDS} \cite{complex2013}}} & `\textit{drive}' & 0.36 & \textbf{0.90} & \textbf{0.72} & 0.63 \\ \cline{2-6} 
& `\textit{forest}'         & \textbf{0.81} & 0.77          & 0.79           & \textbf{0.82}  \\ \cline{2-6} 
& `\textit{parking}'        & 0.897         & 0.89          & \textbf{0.902} & \textbf{0.92}  \\ \cline{2-6} 
& `\textit{store}'          & 0.91          & 0.78          & \textbf{0.94}  & \textbf{0.84}  \\ \cline{2-6} 
& `\textit{traffic}'        & 0.73          & 0.83          & \textbf{0.83}  & \textbf{0.85}  \\ \hline
\multirow{9}{*}{\rotatebox[origin=c]{90}{\textbf{CADS} \cite{movingECCV16}}} & `\textit{chameleon}' & \textbf{0.96} & 0.53 & 0.94 & \textbf{0.66} \\ \cline{2-6} 
& `\textit{frog}'           & 0.49          & 0.38          & \textbf{0.50}  & \textbf{0.40}  \\ \cline{2-6} 
& `\textit{glowwormbeetle}' & 0.84          & 0.88          & \textbf{0.92}  & \textbf{0.94}  \\ \cline{2-6} 
& `\textit{scorpion1}'      & \textbf{0.46} & 0.08          & 0.24           & \textbf{0.22}  \\ \cline{2-6} 
& `\textit{scorpion2}'      & 0.58          & 0.48          & \textbf{0.66}  & \textbf{0.56}  \\ \cline{2-6} 
& `\textit{scorpion3}'      & 0.83          & \textbf{0.41} & \textbf{0.84}  & 0.31           \\ \cline{2-6} 
& `\textit{scorpion4}'      & 0.66          & \textbf{0.76} & \textbf{0.79}  & 0.74           \\ \cline{2-6} 
& `\textit{snail}'          & 0.95          & \textbf{0.90} & \textbf{0.98}  & 0.86           \\ \cline{2-6} 
& `\textit{stickinsect}'    & 0.06          & 0.17          & \textbf{0.20}  & \textbf{0.31}  \\ \hline
\end{tabular}
\caption{Comparison of precision and recall on datasets \textbf{CBDS} and \textbf{CADS} with Bideau \cite{movingECCV16}.}
\label{table:eccv}
\vspace{-0.5cm}
\end{table}

%===========================================================

\begin{figure*}[!htb]
\centering
\setlength{\tabcolsep}{1.0pt}
\begin{tabular}{@{} ccccc @{}}
    (a) & 	
    \begin{subfigure}{.235\textwidth} 
	    \centering 
	    \includegraphics[width=1.0\linewidth]{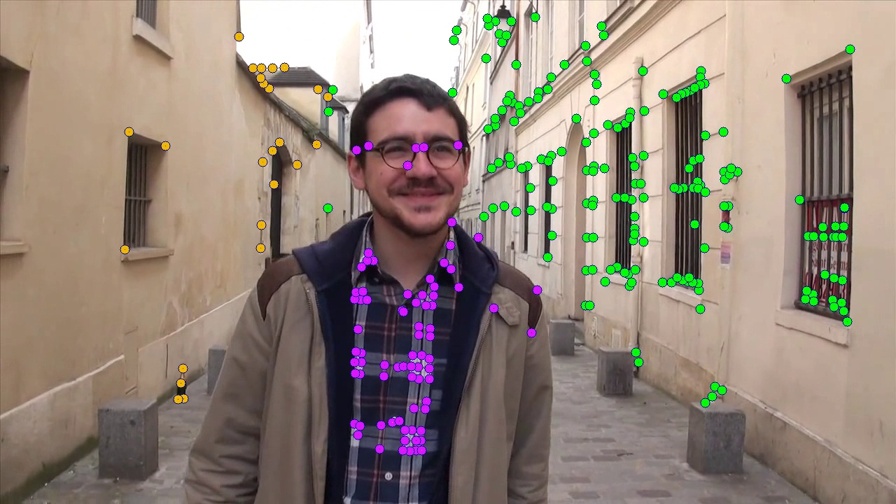} 
	    \vspace{-0.35cm} 
	\end{subfigure} &
	\begin{subfigure}{.235\textwidth} 
	    \centering 
	    \includegraphics[width=1.0\linewidth]{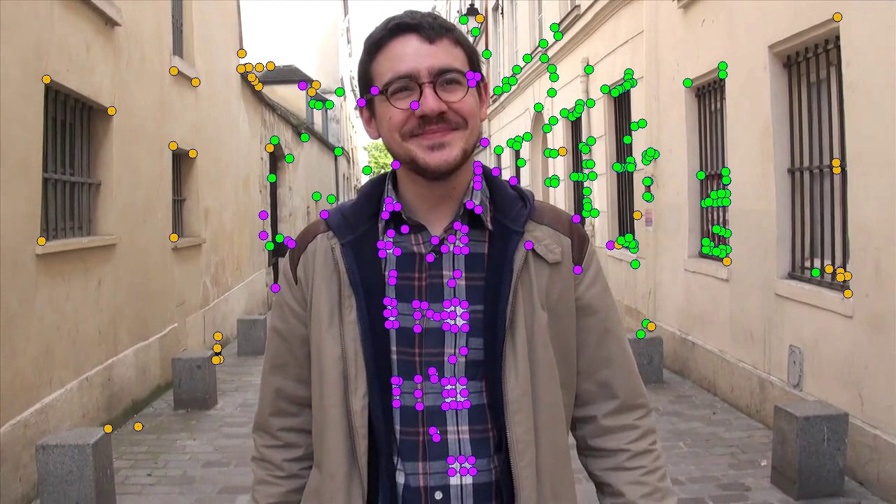} 
	    \vspace{-0.35cm} 
	\end{subfigure} &
    \begin{subfigure}{.235\textwidth} 
	    \centering 
	    \includegraphics[width=1.0\linewidth]{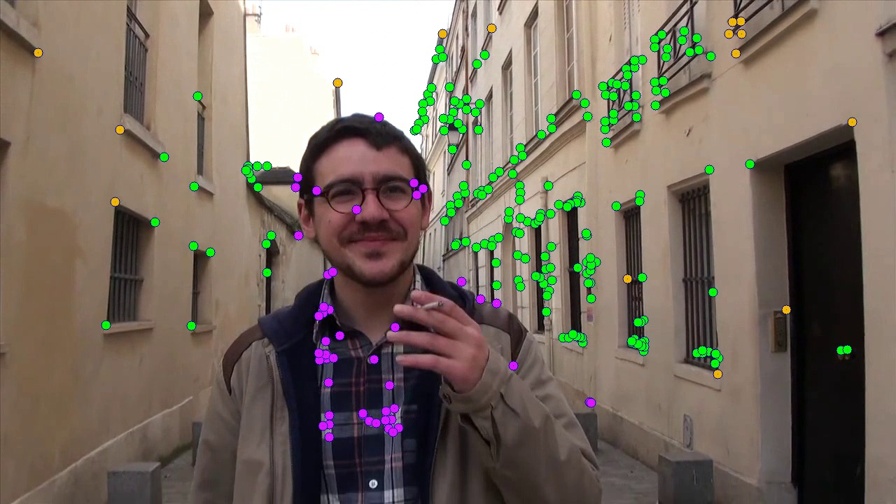} 
	    \vspace{-0.35cm} 
	\end{subfigure} &
    \begin{subfigure}{.235\textwidth} 
	    \centering 
	    \includegraphics[width=1.0\linewidth]{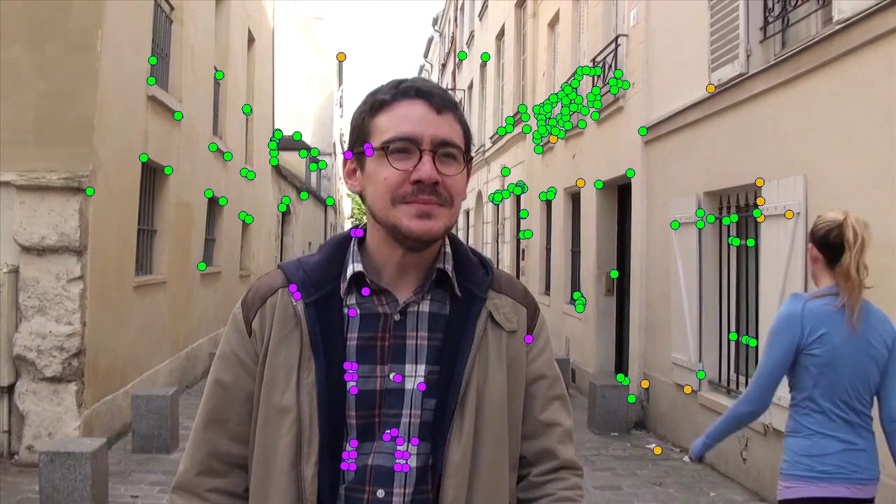} 
	    \vspace{-0.35cm} 
	\end{subfigure}
\\
    (b) & 	
    \begin{subfigure}{.235\textwidth} 
	    \centering 
	    \includegraphics[width=1.0\linewidth]{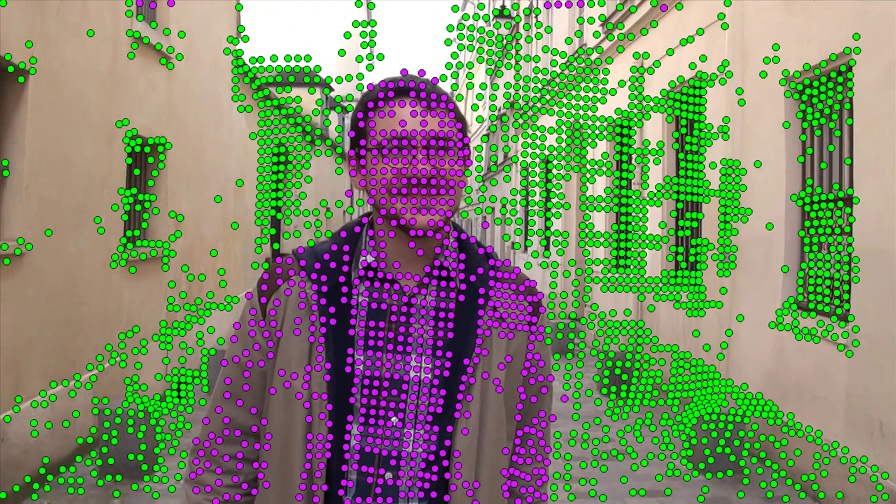} 
	    \vspace{-0.35cm} 
	\end{subfigure} &
	\begin{subfigure}{.235\textwidth} 
	    \centering 
	    \includegraphics[width=1.0\linewidth]{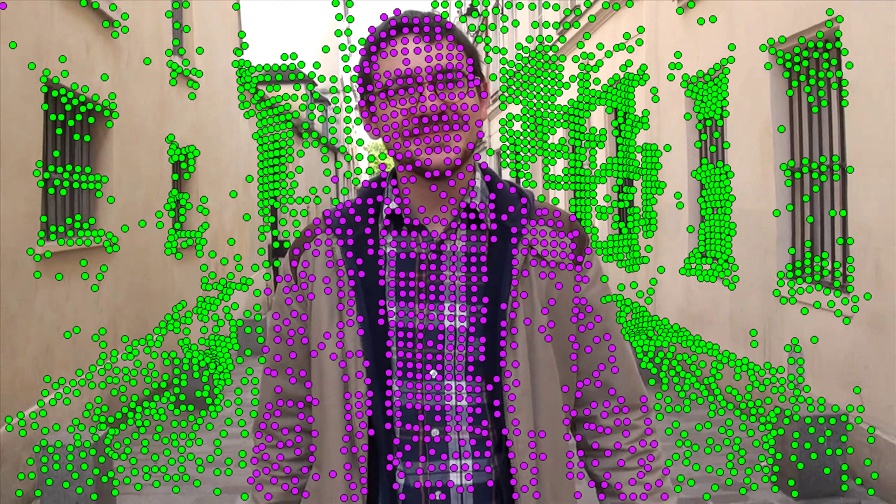} 
	    \vspace{-0.35cm} 
	\end{subfigure} &
    \begin{subfigure}{.235\textwidth} 
	    \centering 
	    \includegraphics[width=1.0\linewidth]{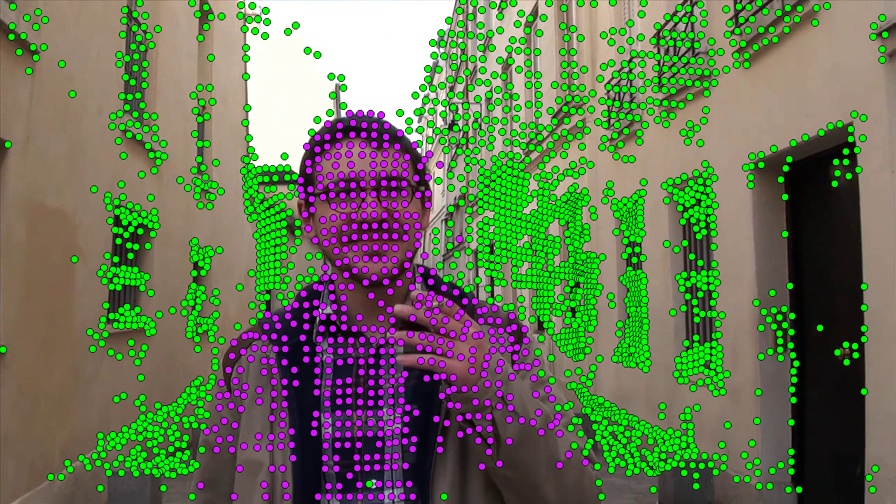} 
	    \vspace{-0.35cm} 
	\end{subfigure} &
    \begin{subfigure}{.235\textwidth} 
	    \centering 
	    \includegraphics[width=1.0\linewidth]{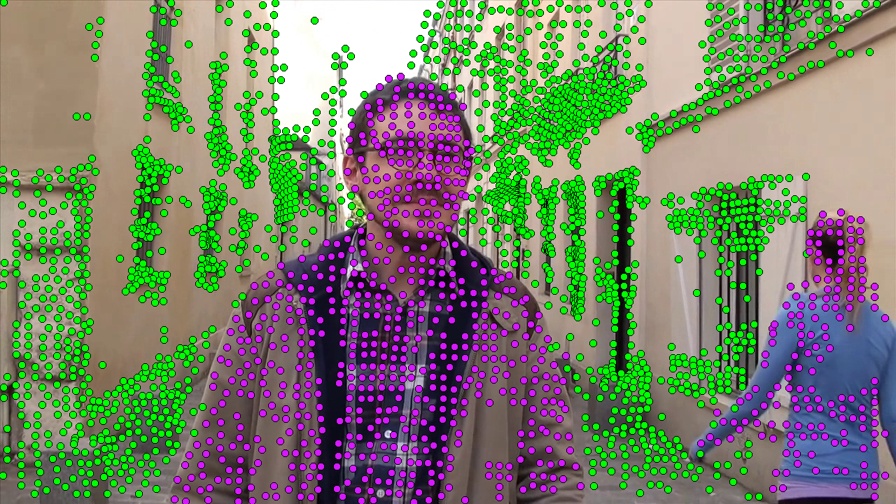} 
	    \vspace{-0.35cm} 
	\end{subfigure}
\\
    (c) & 	
    \begin{subfigure}{.235\textwidth} 
	    \centering 
	    \includegraphics[width=1.0\linewidth]{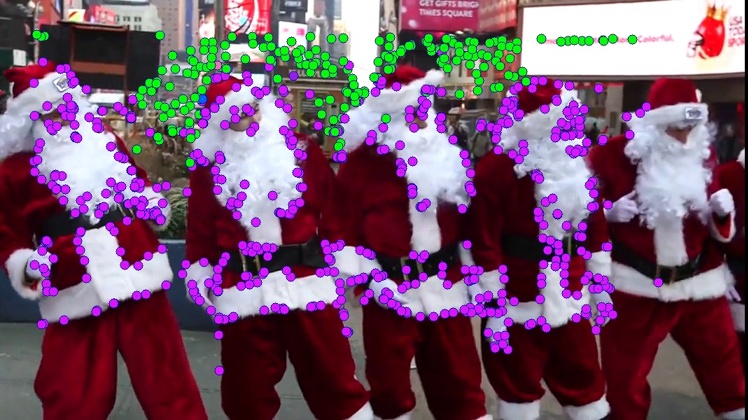} 
	    \vspace{-0.35cm} 
	\end{subfigure} &
	\begin{subfigure}{.235\textwidth} 
	    \centering 
	    \includegraphics[width=1.0\linewidth]{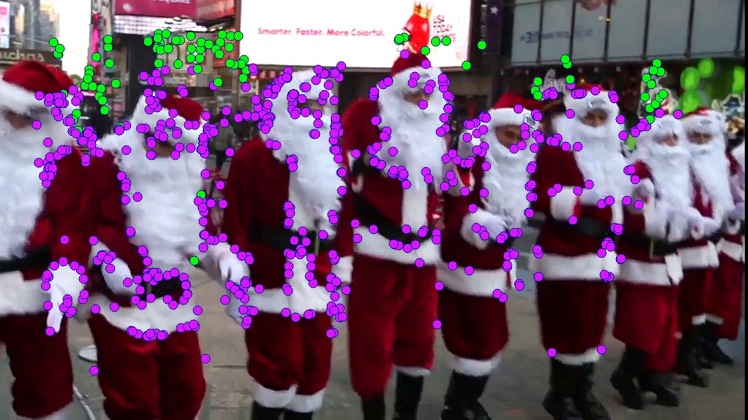} 
	    \vspace{-0.35cm} 
	\end{subfigure} &
    \begin{subfigure}{.235\textwidth} 
	    \centering 
	    \includegraphics[width=1.0\linewidth]{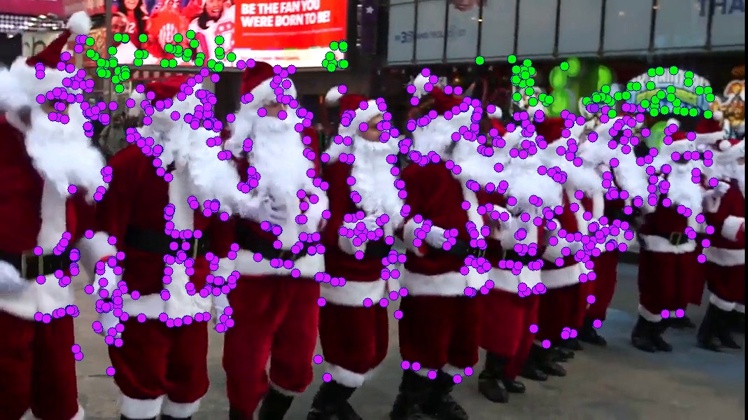} 
	    \vspace{-0.35cm} 
	\end{subfigure} &
    \begin{subfigure}{.235\textwidth} 
	    \centering 
	    \includegraphics[width=1.0\linewidth]{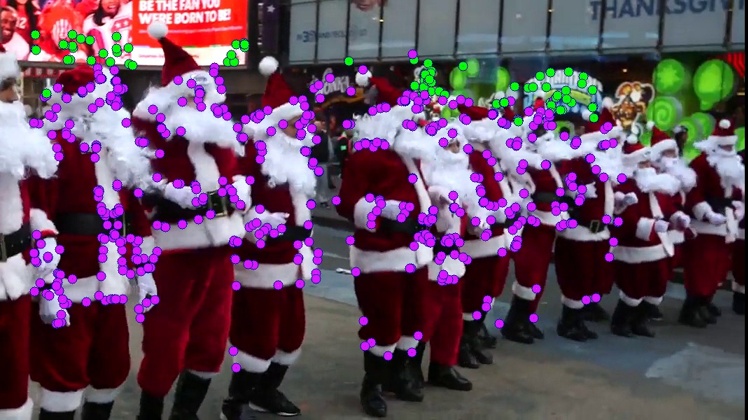} 
	    \vspace{-0.35cm} 
	\end{subfigure}
\\
    (d) & 	
    \begin{subfigure}{.235\textwidth} 
	    \centering 
	    \includegraphics[width=1.0\linewidth]{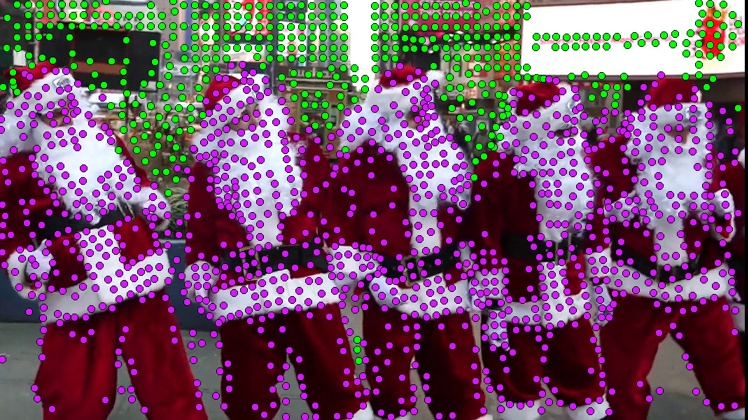} 
	    \vspace{-0.35cm} 
	\end{subfigure} &
	\begin{subfigure}{.235\textwidth} 
	    \centering 
	    \includegraphics[width=1.0\linewidth]{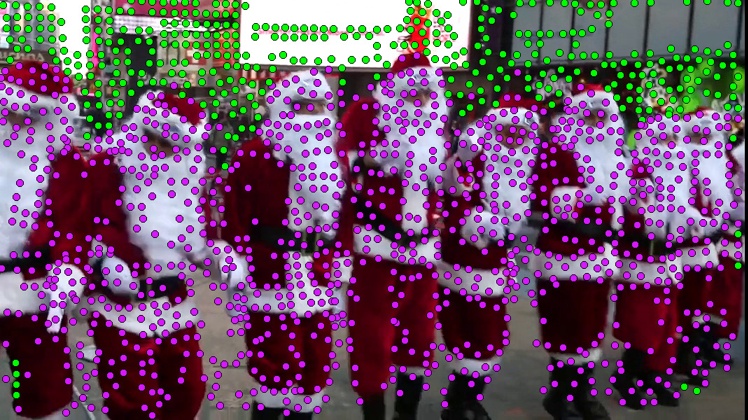} 
	    \vspace{-0.35cm} 
	\end{subfigure} &
    \begin{subfigure}{.235\textwidth} 
	    \centering 
	    \includegraphics[width=1.0\linewidth]{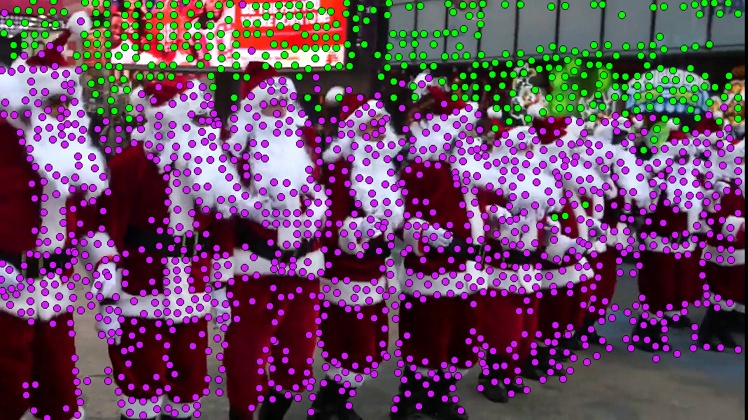} 
	    \vspace{-0.35cm} 
	\end{subfigure} &
    \begin{subfigure}{.235\textwidth} 
	    \centering 
	    \includegraphics[width=1.0\linewidth]{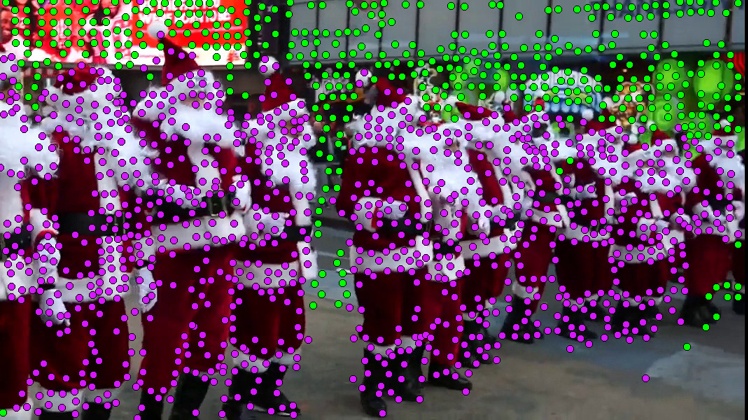} 
	    \vspace{-0.35cm} 
	\end{subfigure}
\\
    (e) & 	
    \begin{subfigure}{.235\textwidth} 
	    \centering 
	    \includegraphics[width=1.0\linewidth]{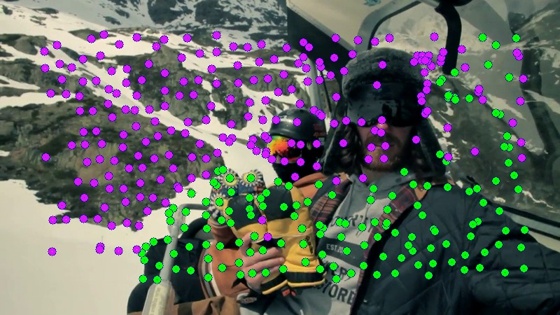} 
	    \vspace{-0.35cm} 
	\end{subfigure} &
	\begin{subfigure}{.235\textwidth} 
	    \centering 
	    \includegraphics[width=1.0\linewidth]{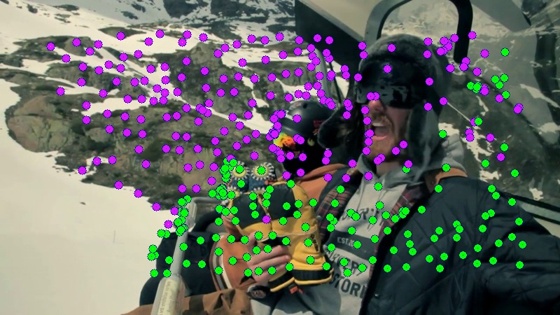} 
	    \vspace{-0.35cm} 
	\end{subfigure} &
    \begin{subfigure}{.235\textwidth} 
	    \centering 
	    \includegraphics[width=1.0\linewidth]{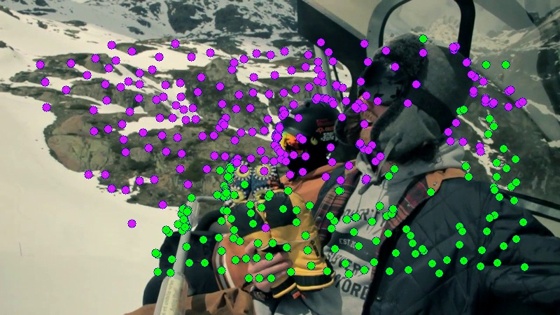} 
	    \vspace{-0.35cm} 
	\end{subfigure} &
    \begin{subfigure}{.235\textwidth} 
	    \centering 
	    \includegraphics[width=1.0\linewidth]{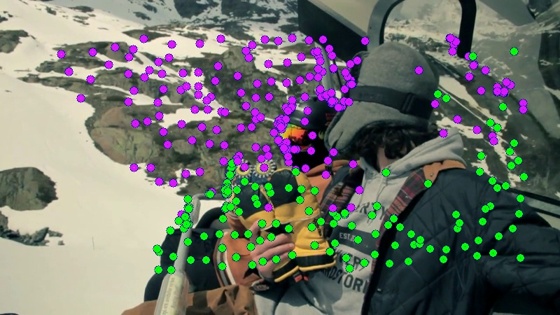} 
	    \vspace{-0.35cm} 
	\end{subfigure}
\\
    (f) & 	
    \begin{subfigure}{.235\textwidth} 
	    \centering 
	    \includegraphics[width=1.0\linewidth]{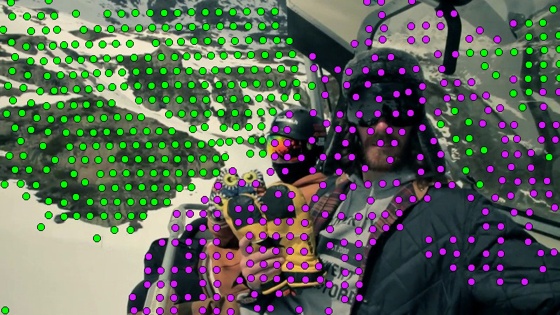} 
	    \vspace{-0.35cm} 
	\end{subfigure} &
	\begin{subfigure}{.235\textwidth} 
	    \centering 
	    \includegraphics[width=1.0\linewidth]{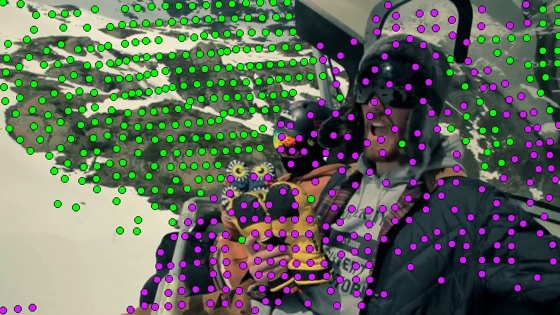} 
	    \vspace{-0.35cm} 
	\end{subfigure} &
    \begin{subfigure}{.235\textwidth} 
	    \centering 
	    \includegraphics[width=1.0\linewidth]{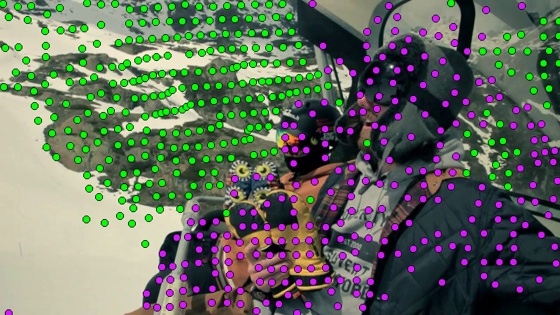} 
	    \vspace{-0.35cm} 
	\end{subfigure} &
    \begin{subfigure}{.235\textwidth} 
	    \centering 
	    \includegraphics[width=1.0\linewidth]{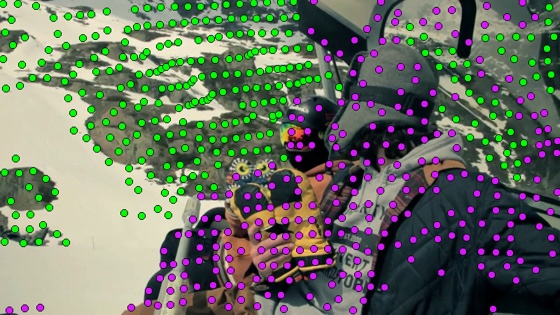} 
	    \vspace{-0.35cm} 
	\end{subfigure}
\\
\end{tabular}
\caption{Comparison with Zhang \textit{et al.} \cite{robust2016}. Row (a), (c), and (e) are results 
from \cite{robust2016}. Row (b), (d), and (f) are our results.}
\label{fig:sig}
\end{figure*}

%===========================================================
\subsection{Limitations and Discussions}
Our method may not work well in some special cases. Firstly, for videos 
with many short feature trajectories (only $2\sim3$ feature points on a 
trajectory) extracted from non-rigid objects like river and sea, where the 
features on these subtle dynamically moving objects may not violate the 
epipolar constraint during their short lifetime. Therefore, they may be 
wrongly labeled as background. Secondly, for videos captured by camera that 
hardly moves, if the foreground motion is also rigid and occupies the 
majority of the scene all the time (\eg, a video recorded by a standing 
person with a close-up view of a running train in front), our method may 
also fail to identify the correct background features. Computational wise, 
our method usually takes less than five minutes to process a video of around 
100 frames and 50,000 trajectories on a PC with 2.4GHz CPU.

%===========================================================
\section{Conclusion}
In this work, we propose a robust background feature identification 
method that can handle moving foreground objects that are large or exhibit 
intermittent motions. Our method is designed based on the assumption that 
the background motion will contain the largest amount of feature trajectories 
when the local background trajectories are aggregated over the entire 
sequence. Accordingly, we develop a local-to-global dominant motion group 
identification pipeline. Since the motions are characterized using 
fundamental matrices, there is no issue with over-segmentation, problem 
that plagues the 2D motion segmentation approach. With careful design, 
our motion estimation method can efficiently handle large amount of 
trajectories and robustly propose potential rigid motions in video clips. 
The comprehensive experiment results show that our method outperforms 
several most recent state-of-the-art motion segmentation methods both 
quantitatively and qualitatively. 

\bibliographystyle{splncs04}
\bibliography{egbib}

\end{document}